\documentclass[letterpaper, 10 pt, conference]{ieeeconf}  

\IEEEoverridecommandlockouts                              

\newcommand{\projectname}{MobiDock}

\title{\LARGE \bf
\projectname{}: Design and Control of A Modular Self Reconfigurable\\Bimanual Mobile Manipulator via Robotic Docking
}

\author{
Xuan Thuan Nguyen, Khac Nam Nguyen, Ngoc Duy Tran, Thi Thoa Mac, Anh Nguyen,\\ Hoang Hiep Ly, Tung D. Ta*
}

\usepackage{amsmath}
\usepackage{graphicx}
\usepackage{subcaption} 
\usepackage{flushend}
\usepackage[font=small]{caption}   
\usepackage{balance}
\usepackage{float}
\usepackage{siunitx}
\usepackage{tabularx}
\usepackage{booktabs}
\usepackage{color, colortbl}
\usepackage[maxfloats=256]{morefloats}
\makeatletter
\let\NAT@parse\undefined
\makeatother
\usepackage[bookmarks=false, linkcolor=blue, urlcolor=blue, citecolor=blue]{hyperref} 
\usepackage{fancyhdr}
\hypersetup{
    colorlinks=true,
    linkcolor=red,
    filecolor=magenta,      
    urlcolor=blue,
    pdfstartview={FitH},
    citecolor =blue
    }
    
\begin{document}

\maketitle
\thispagestyle{empty}
\pagestyle{empty}

\begin{abstract}

Multi-robot systems, particularly mobile manipulators, face inherent challenges in coordination and dynamic stability during collaborative tasks. Rather than focusing on complex software-based coordination algorithms, this study proposes \projectname{}, a modular reconfigurable system that introduces a physical docking mode as a structural solution to these challenges. By allowing two independent mobile manipulators to physically connect via a vision-based autonomous docking strategy and a threaded screw-lock mechanism, the system transforms into a unified bimanual platform. This hardware-level reconfiguration simplifies the control problem by treating the multi-robot assembly as a single kinematic entity, thereby augmenting the operational range of multi-robot teams. Experimental results demonstrate that this additional docked mode significantly outperforms traditional independent cooperation in dynamic stability and efficiency. Specifically, the unified configuration achieves lower Root Mean Square Acceleration (RMSA) and Jerk values, better angular precision, and faster task completion times. These findings confirm that integrating a physical docking mode is a powerful design principle that complements existing multi-robot systems, offering a highly stable and efficient alternative for complex tasks in real-world environments.

\end{abstract}

\section{INTRODUCTION}

Robotics is rapidly evolving toward highly flexible, adaptable systems. One promising direction is the development of reconfigurable robots that can change their morphology to match different tasks and environments~\cite{Yim2007}. This ability to adapt both form and function is a major advantage, improving robustness and versatility in situations where fixed-form robots are less effective. They have been applied across diverse application domains, including industrial manipulation~\cite{Mohamed2015}, autonomous cleaning robot~\cite{Veerajagadheswar2022}, search and rescue operation~\cite{Cordie2019}, and space exploration~\cite{Li2025}.

Among reconfigurable robot architectures, modular self-reconfigurable robots (MSRRs) are an increasingly prominent and fast-growing direction. These systems include multiple independent robotic modules that can be interconnected to form various configurations~\cite{Liang2023},~\cite{Dokuyucu2023}. This modularity offers several advantages, including increased flexibility and the ability to overcome the physical limitations of a single module, such as restricted reach or payload capacity~\cite{Li2023},~\cite{Walker2024}. MSRRs are particularly valuable for tasks that require a change in form to navigate different environments, for example, SolderCubes~\cite{Neubert2016}, SMORES-EP~\cite{Liu2023}, RS-ModCubes~\cite{Zheng2025}, M-TRAN~\cite{Murata2002}, and FireAnt~\cite{Swissler2020} demonstrating these capabilities in many applications.

\begin{figure}[h!]
\centering
\includegraphics[width=\linewidth]{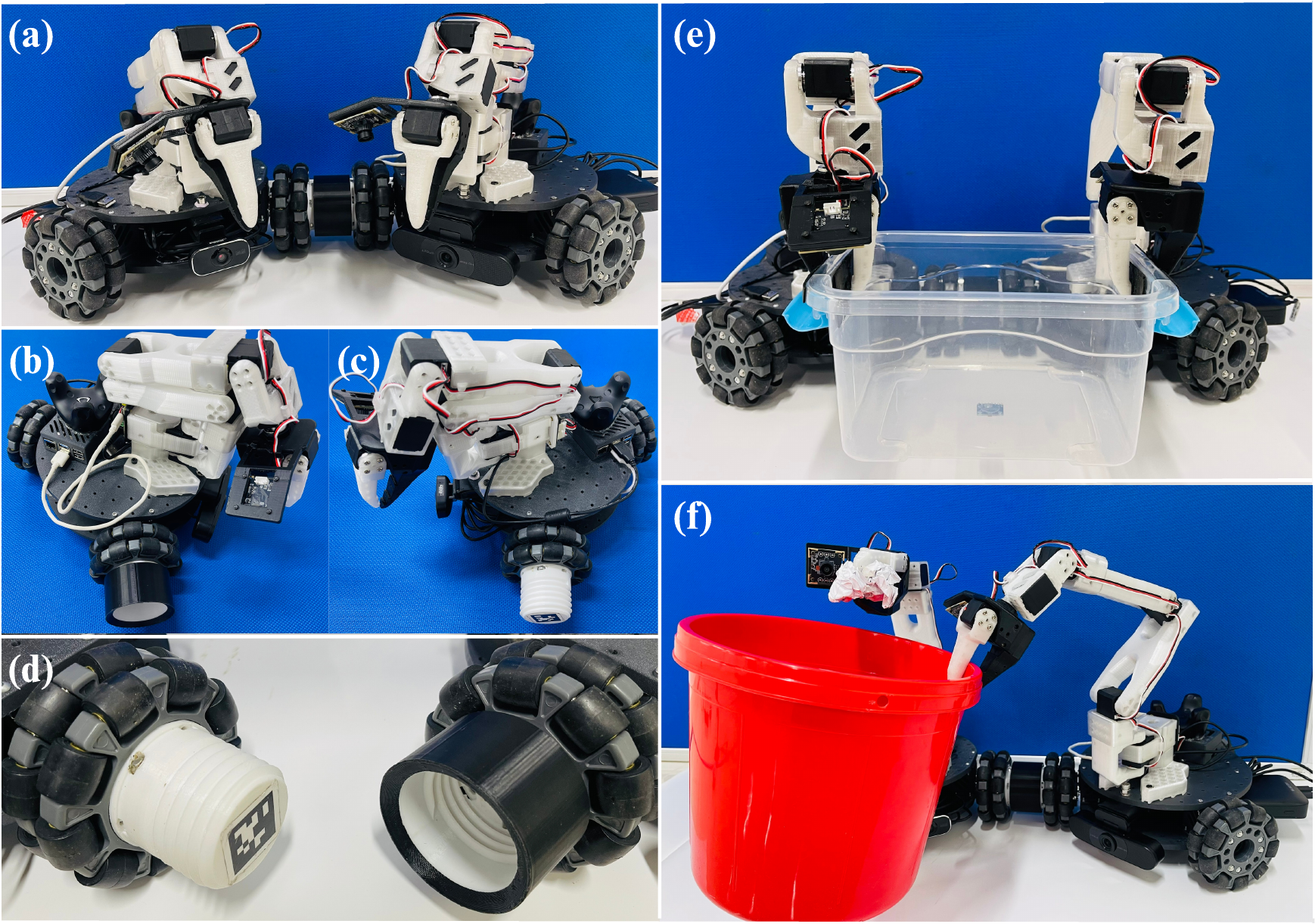}
\caption{Key components of the proposed \projectname{} system and its experimental applications: (a)~the post-reconfigured bimanual robot system, (b-c)~the independent mobile manipulator modules before docking, (d)~the mechanical docking mechanism, (e)~the system undergoes a stability test while lifting a box, (f)~a cooperative manipulation experiment demonstrates the system's performance by picking up trash and putting it in a bin.}
\label{fig:overal}
\vspace{-6mm}
\end{figure}

Besides the development of MSRRs, mobile manipulator robots have long been recognized for their dexterity and flexibility, combining the mobility of a wheeled base with the precision of a robotic arm~\cite{Pu2023},~\cite{Yu2018}. This integration has opened new possibilities for tackling complex tasks in unstructured environments~\cite{Ruchika2024},~\cite{Zhang2025},~\cite{Lian2024}.  However, flexibility comes with a considerable increase in control complexity and a lack of dynamic stability, particularly when two or more mobile manipulators must coordinate to perform a shared task~\cite{Sun2023},~\cite{Zhai2022},~\cite{Chen2018}. The main challenge is the strong kinematic and dynamic coupling between robots, which demands advanced control strategies to ensure formation stability, collision avoidance, and minimal internal forces.
Several approaches have been proposed to address this challenge, including distributed control, adaptive algorithms, and reinforcement learning, each demonstrating encouraging results.~\cite{Ren2020},~\cite{Wu2023},~\cite{Marino2018},~\cite{Xu2023}. In practice, these approaches typically rely on high communication bandwidth, fast computation, and accurate state estimation, requirements that are challenging to meet in chaotic real-world environments. Consequently, many solutions perform well in laboratory settings but remain difficult to scale in real-world deployment.

This study addresses a critical gap: while MSRRs provide high structural adaptability and mobile manipulators offer superior dexterity, a system that seamlessly transitions between independent, cooperative, and unified states remains an open challenge. We propose \projectname{}, a modular reconfigurable mobile manipulator system designed to augment the operational repertoire of multi-robot teams by introducing a high-stability docking mode. The central concept of \projectname{} is to allow independent mobile manipulators to physically dock, transforming a complex multi-robot coordination problem into the control of a single, unified bimanual platform. Crucially, this approach is not intended to replace traditional software-based cooperative strategies; rather, it provides an additional, hardware-enabled operational mode specifically optimized for tasks requiring high load capacity and dynamic stability. By establishing a rigid physical link, we effectively ``collapse'' the coordination complexity, bypassing the synchronization bottlenecks and communication latencies inherent in multi-robot teams while preserving the individual flexibility of each module.

The primary contributions of this paper, visually summarized in Fig.~\ref{fig:overal}, are as follows:
\begin{enumerate}
\item The design and realization of \textbf{\projectname{}}, a system that enables a multi-modal operational framework, allowing robots to switch between separated mobile manipulators for independent tasks execution and a unified, rigid bi-manual configuration.
\item An integrated control strategy that manages the entire lifecycle of reconfiguration, including a robust vision-based docking procedure and a unified control law for the post-docking phase.
\item A comprehensive experimental validation demonstrating that the proposed docking mode provides a superior alternative for stability-critical tasks compared to traditional independent coordination.
\end{enumerate}

\section{Preliminaries}
\subsection{System Overview}

This study introduces a reconfigurable mobile manipulator system designed to dynamically form new configurations by docking multiple modules. This allows the system to adjust its kinematic structure and payload capacity for tasks ranging from cooperative transport to complex multi-arm manipulation, providing a flexible solution to limitations inherent in fixed-configuration robots.

The system's basic module, based on the LeKiwi\footnote{\url{https://huggingface.co/docs/lerobot/lekiwi}} design, features a 6-DOF manipulator mounted on a three-wheel omnidirectional mobile base. This triangular omni-wheel configuration provides the maneuverability required for precise docking, allowing for lateral and rotational adjustments without complex maneuvers while the manipulator executes diverse tasks.

Integrating modules via a base-level coupling mechanism presents a distinct control challenge compared to arm-based docking. However, the omnidirectional platform mitigates this by enabling precise pose attainment and compensation for external disturbances. Furthermore, this design simplifies the kinematic control of the reconfigured system by decoupling translational and rotational motions, ensuring an efficient transition into a unified multi-arm platform.

\subsection{Kinematic Modeling}
\begin{figure}[h!]
\centering
\vspace{-0.3cm}
\includegraphics[width=0.8\linewidth]{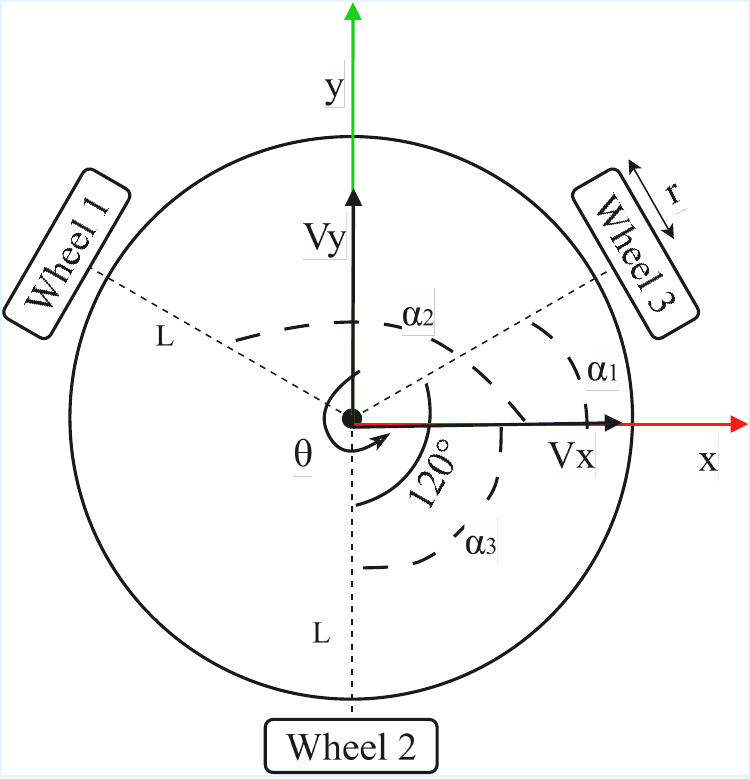}
\vspace{-0.0cm}
    \caption{A 3-wheeled omnidirectional robot model.}
\vspace{-0.2cm}
\label{fig:kinematic_model}

\end{figure}

The mobile base consists of a rigid body with three omnidirectional wheels, each placed at a \SI{120}{\degree} interval from the others. We define the robot's pose in the global frame as a vector $\mathbf{q} = [x, y, \theta]^T$, where $(x, y)$ are the coordinates of the robot's center and $\theta$ is its orientation. The robot's linear and angular velocities are represented by $\mathbf{\dot{q}} = [\dot{x}, \dot{y}, \dot{\theta}]^T$. As shown in Fig.~\ref{fig:kinematic_model}, each wheel, with a radius $r$, has a rotational velocity of $\dot{\phi}_i$ where $i \in \{1, 2, 3\}$.

The relationship between the robot's velocity and the rotational velocities of its wheels is given by the following equation:

\begin{equation}
\begin{bmatrix}
\dot{\phi}_1 \\
\dot{\phi}_2 \\
\dot{\phi}_3
\end{bmatrix}
=
\frac{1}{r}
\begin{bmatrix}
-\sin(\alpha_1) & \cos(\alpha_1) & L \\
-\sin(\alpha_2) & \cos(\alpha_2) & L \\
-\sin(\alpha_3) & \cos(\alpha_3) & L
\end{bmatrix}
\begin{bmatrix}
\dot{x} \\
\dot{y} \\
\dot{\theta}
\end{bmatrix}
\label{equa:kinematic}
\end{equation}
where $L$ is the distance from the robot's central point to the center of each wheel. The mounting angles of the three wheels are defined as $\alpha_1, \alpha_2,$ and $\alpha_3$, which are set to \SI{150}{\degree}, \SI{-90}{\degree} and \SI{30}{\degree} respectively in our system.

\section{Reconfigurable Mobile Manipulator System}
To enhance the versatility of robotic systems in dynamic environments, we propose a reconfigurable mobile manipulator capable of adapting its morphology to task-specific requirements. This modular approach allows individual units to dock physically, creating a unified platform with augmented capabilities. The following subsections detail the system's core components: \textbf{(A)} Mechanical Design, \textbf{(B)} Docking Strategy, and \textbf{(C)} Post-Reconfigured System Analysis. 
\subsection{Mechanical Design}

This section presents the mechanical design of the modular docking mechanism, integrated directly into the side of each module's omnidirectional wheel. To ensure a compact and reliable connection, the design meets three core requirements: (i) independence from wheel rotation, (ii) utilization of existing wheel motor torque for engagement, and (iii) provision of a flat surface for visual perception.

\begin{figure}[h!]
\centering
\includegraphics[width=\linewidth]{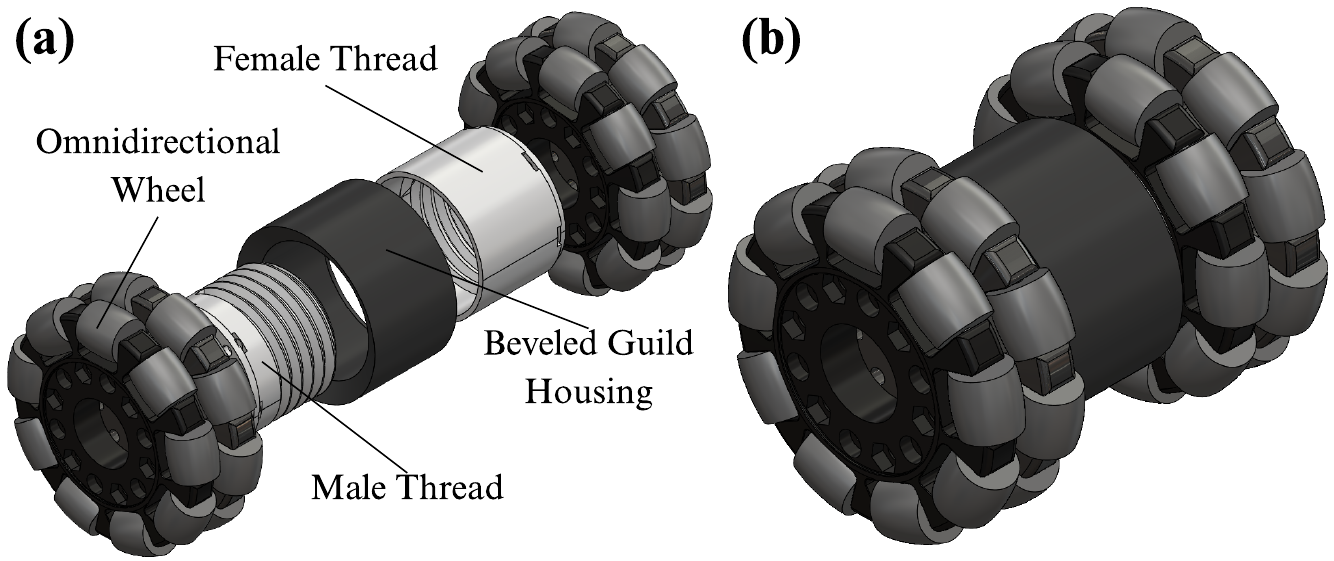}
\caption{The key components and a final configuration of the docking mechanism: (a)~exploded view of the proposed screw-based docking mechanical design. The docking hubs are attached to one of the wheels of the two robots. During the docking process, the female hub remains stationary while the male hub rotates to fasten the docking. (b)~Overview of the two modules fully docked, forming a single, rigid reconfigured system.}
\vspace{-0.5cm}
\label{fig:mechanicaldesign}
\end{figure}

Based on these criteria, we propose a threaded screw-lock mechanism (specifications in Table~\ref{tab:mechanicalparam_new}). The design features a large triangular cross-section and beveled leading edges (Fig.~\ref{fig:mechanicaldesign}a) to maximize the capture envelope and facilitate self-alignment.

The proposed screw-lock mechanism provides a strategic balance between structural rigidity and power efficiency. Unlike phase-change systems such as SolderCubes \cite{Neubert2016}, which require high thermal power ($\approx$15W) and long operational cycles ($\approx$30s), \projectname{} minimizes energy overhead. By leveraging an "actuator-sharing" strategy with locomotion motors, the system eliminates auxiliary actuators while its self-locking geometry ensures zero power consumption in the holding state.

As summarized in Table~\ref{tab:docking_compare}, our threaded engagement generates substantial axial clamping force (pre-loading), outperforming magnetic systems (e.g., SMORES-EP~\cite{Liu2023}) that lack shear resistance, or mechanical latches (e.g., FireAnt~\cite{Swissler2020}) prone to backlash. This pre-loading effect ensures a rigid, single-body configuration (Fig.~\ref{fig:mechanicaldesign}b) and a higher alignment tolerance (±5 mm) than most mechanical alternatives. Thus, \projectname{} represents a power-autonomous and robust solution for heavy-duty modular manipulation.


\begin{table}[h]
\centering
\caption{Technical specifications of the proposed mechanical design}
\label{tab:mechanicalparam_new}
\begin{tabular}{ll}
\toprule
\textbf{Parameter} & \textbf{Value} \\
\midrule
Thread type & ISO metric thread \\
Designation & $\text{M}56 \times \SI{5.5}{}$ \\
Nominal diameter, $d$ & \SI{56}{\milli\meter} \\
Pitch, $P$ & \SI{5.5}{\milli\meter} \\
Major diameter, $d_{max}$ & \SI{56.000}{\milli\meter} \\
Pitch diameter, $d_2$ & \SI{53.917}{\milli\meter} \\
Minor diameter, $d_1$ & \SI{51.835}{\milli\meter} \\
Theoretical thread height, $H$ & \SI{4.763}{\milli\meter} \\
Effective thread height, $h$ & \SI{3.372}{\milli\meter} \\
Flank angle & \SI{60}{\degree} \\
Tolerance class & 6g/6H \\
\bottomrule
\end{tabular}
\vspace{-0.15cm}
\end{table}

\begin{table*}[h]
\centering
\caption{Comparison of different docking mechanisms}
\label{tab:docking_compare}
\begin{tabular}{lllllrr}
\toprule
\textbf{Docking design} &
\textbf{Representative} &
\textbf{Holding force} &
\textbf{Auxiliary Actuators} &
\textbf{Holding Power} &
\textbf{Alignment Tolerance} &
\textbf{Docking time} \\
\midrule

Phase-change     & SolderCubes~\cite{Neubert2016}      & High   & 1 (Heater)& No  & $\pm$2 mm    & $\approx$30s \\
Magnetic         & SMORES-EP~\cite{Liu2023}       & Low    & 0         & Yes & $\pm$10 mm  & $\approx$1s \\
Mechanical latch & FireAnt~\cite{Swissler2020}, M-TRAN~\cite{Murata2002} & Medium & 1 (Motor) & No  & $\pm$2 mm    & $\approx$5s  \\
\textbf{Screw-lock}       & \textbf{\projectname{} (Ours)}        & \textbf{Medium} & \textbf{0}         & \textbf{No}  & \textbf{$\pm$5 mm} & \textbf{$\approx$15s} \\

\bottomrule
\end{tabular}
\vspace{-0.15cm}
\end{table*}


\subsection{Docking Strategies} 

\begin{figure}[h!]
\centering
\includegraphics[width=\linewidth]{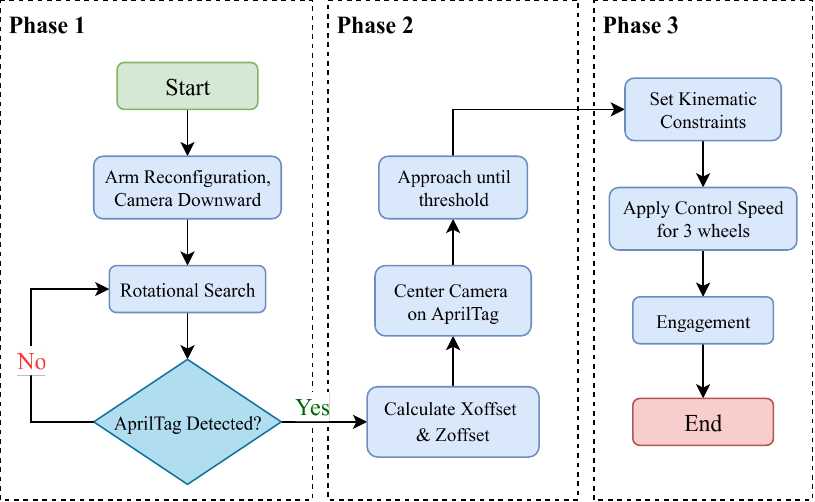}
\caption{A schematic overview of the reconfiguration procedure between robotic modules, highlighting the main stages leading to physical connection.}
\vspace{-0.5cm}
\label{fig:diagram}
\end{figure}

Successful physical connection requires a robust vision-based strategy to detect and engage the docking mechanisms accurately. We utilize AprilTag markers~\cite{olson2011apriltag} and an arm-mounted camera to execute a three-phase maneuver (Fig.~\ref{fig:diagram}).

\subsubsection{Phase 1}

\begin{figure}[h!]
\centering
\vspace{-0.2cm}
\includegraphics[width=0.48\textwidth]{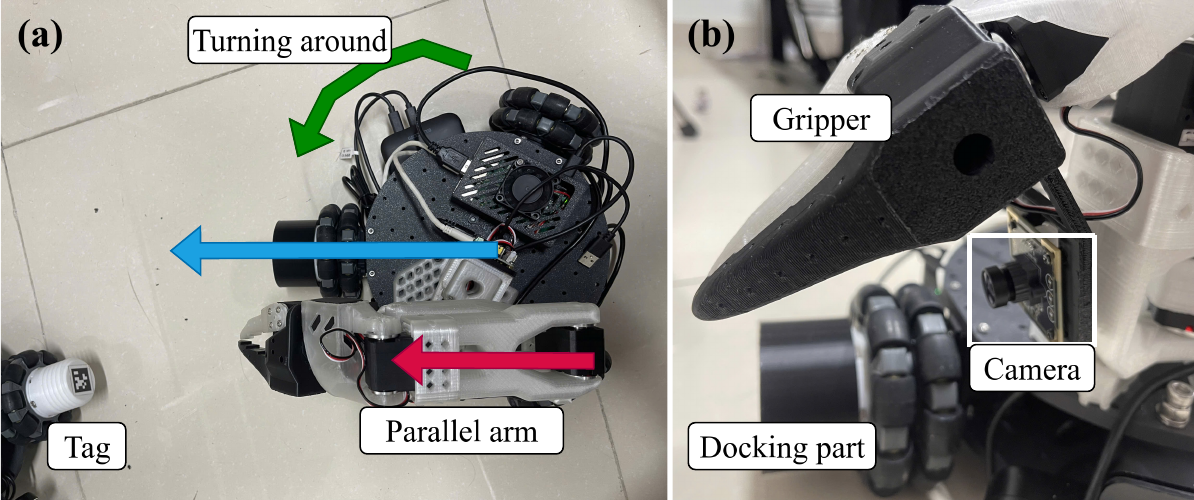}
\caption{The initial operational phase of a \projectname{} module, (a) the robot rotates to find the target tag, with its manipulator parallel to the docking mechanism, (b) the camera is then rotated downward to precisely detect and localize the tag.}
\label{fig:phase1}
\vspace{-0.2cm}
\end{figure}

The system first reconfigures the manipulator arm to align with the docking wheel while directing the camera downward. The mobile base performs a rotational scan to detect the target AprilTag within the surrounding environment (Fig.~\ref{fig:phase1}).

\subsubsection{Phase 2}

\begin{figure}[H]
\centering
\includegraphics[width=\linewidth]{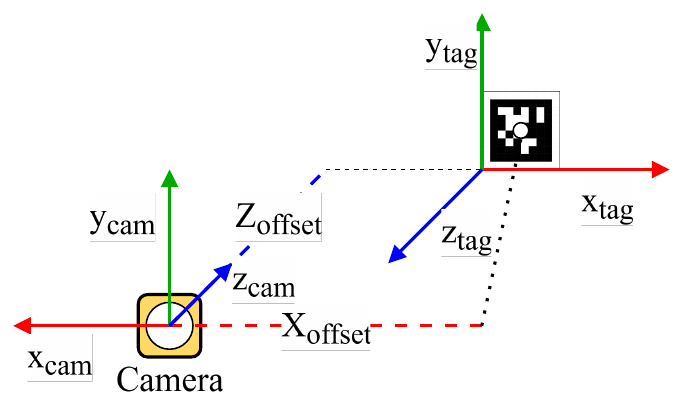}
\caption{Illustrations of AprilTag and camera axes. Lateral $X_\text{offset}$ and depth $Z_\text{offset}$ offsets guide the robot to center the camera on the tag, ensuring precise alignment during approach.}
\vspace{-0.5cm}
\label{fig:apriltag}
\end{figure}

Upon detection, the robot centers the camera on the tag and approaches until a predefined depth is reached. Precision alignment is maintained by minimizing two key parameters: the lateral offset $X_\text{offset}$ and depth offset $Z_\text{offset}$, guiding the robot into the final docking pose (Fig.~\ref{fig:apriltag}).

\subsubsection{Phase 3}

\begin{figure}[h!]
\centering
\vspace{-0.5cm}
\includegraphics[width=0.8\linewidth]{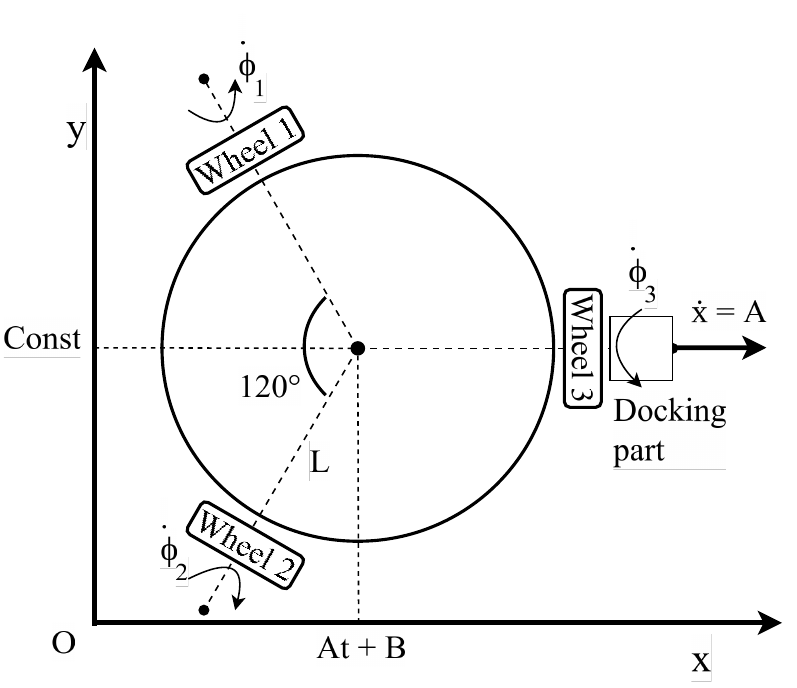}
\caption{Visualization of the kinematic law for locking, where zero lateral motion and controlled forward/rotational velocities coordinate wheel motion to achieve proper thread engagement during docking.}
\vspace{-0.4cm}
\label{fig:phase3}
\end{figure}

The final phase executes the locking maneuver by combining forward motion with controlled torsional force to engage the threads. To counteract the omnidirectional base's tendency to follow an arcing path—which hinders engagement—we implemented a kinematic control law with specific constraints: zero lateral velocity $(\dot{y}=0)$, constant forward velocity $(\dot{x}= \text{A})$, and a constant rotational velocity for the docking wheel $(\dot{\phi}_3)$ to provide necessary screwing torque. Based on the kinematics in (\ref{equa:kinematic}), the velocities of the remaining wheels are calculated as:

\begin{equation}
\begin{bmatrix}
\dot{\phi}_1 \\
\dot{\phi}_2
\end{bmatrix}
=
\frac{1}{r}
\begin{bmatrix}
\cos(\alpha_1) & L \\
\cos(\alpha_2) & L
\end{bmatrix}
\begin{bmatrix}
\dot{x} \\
\dot{\theta}
\end{bmatrix}
\end{equation}

where $\dot{\theta} = \dot{{\phi_3}}$ (the constant angular velocity of the docking wheel). The motion plan is visually described in Fig.~\ref{fig:phase3}. Based on this control method, we can ensure both the necessary translational force for a secure press and a synchronized rotational torque for thread engagement. This approach facilitates the robust and seamless locking of the threaded mechanism, thereby ensuring a successful docking process.

\subsection{Post-Reconfigured System Analysis}
\label{sec:3C}

The modularity of our design ensures that fundamental physical principles remain applicable to any $n$-module docked system ($n \geq 2$). Upon docking, the unified structure possesses a total mass $M$ and a global center of mass $\mathbf{r}_{\text{CM}}$ determined by:
\begin{equation}
\mathbf{r}_{\text{CM}} = \frac{\sum_{i=1}^{n} m_i \mathbf{r}_i}{\sum_{i=1}^{n} m_i}
\end{equation}
where $m_i$ and $\mathbf{r}_i$ are the mass and center of mass vector of the $i$-th module, respectively. This redistribution ensures the center of mass shifts to maintain balance along the primary axis of the multi-module chain.

Similarly, the system's total moment of inertia $I_{\text{total}}$ is calculated via the Parallel Axis Theorem:
\begin{equation}
I_{\text{total}} = \sum_{i=1}^{n} (I_i + m_i d_i^2)
\end{equation}
where $I_i$ is the individual moment of inertia and $d_i$ is the distance to the new collective center of mass. The cumulative increase in $I_{\text{total}}$, alongside an expanded stability polygon, significantly enhances resistance to external disturbances and tilting moments during heavy payload manipulation.

Beyond these physical shifts, reconfiguration achieves structural rigidity by reducing redundant degrees of freedom (DOF). While $n$ independent modules possess $9n$ DOF, the docking mechanism imposes kinematic constraints that merge the individual bases into a single rigid platform. Consequently, the system’s total DOF is simplified to:
\begin{equation}
\mathrm{DOF}_{\text{total}}
=
\mathrm{DOF}_{\text{base}}
+
\sum_{i=1}^{n} \mathrm{DOF}_{\text{arm},i}
=
3 + 6n
\end{equation}

Eliminating $3(n - 1)$ redundant base DOFs through rigid coupling transforms the independent modules into a unified $x,y,\theta$ platform with enhanced multi-arm dexterity. Unlike software-centric coordination for closed-chain constraints \cite{Marino2018, Xu2023}, \projectname{}’s hardware-level approach collapses the high-dimensional state space into a single centralized model. This bypasses complex internal force modeling and high-frequency peer-to-peer synchronization, significantly reducing both computational overhead and communication bandwidth requirements.

To coordinate translational maneuvers for the n-modules system, a common system axis is defined parallel or perpendicular to a radial wheel vector (as illustrated in Fig.~\ref{fig:Axes}). Each module $i$ synchronizes its heading by applying a rotation matrix $R(\beta_i)$ to its local frame, ensuring all modules move in unison as a single rigid platform. The resulting wheel velocity mapping for the $i$-th module is expressed as:

\begin{figure*}[h!]
\centering
\includegraphics[width=\linewidth]{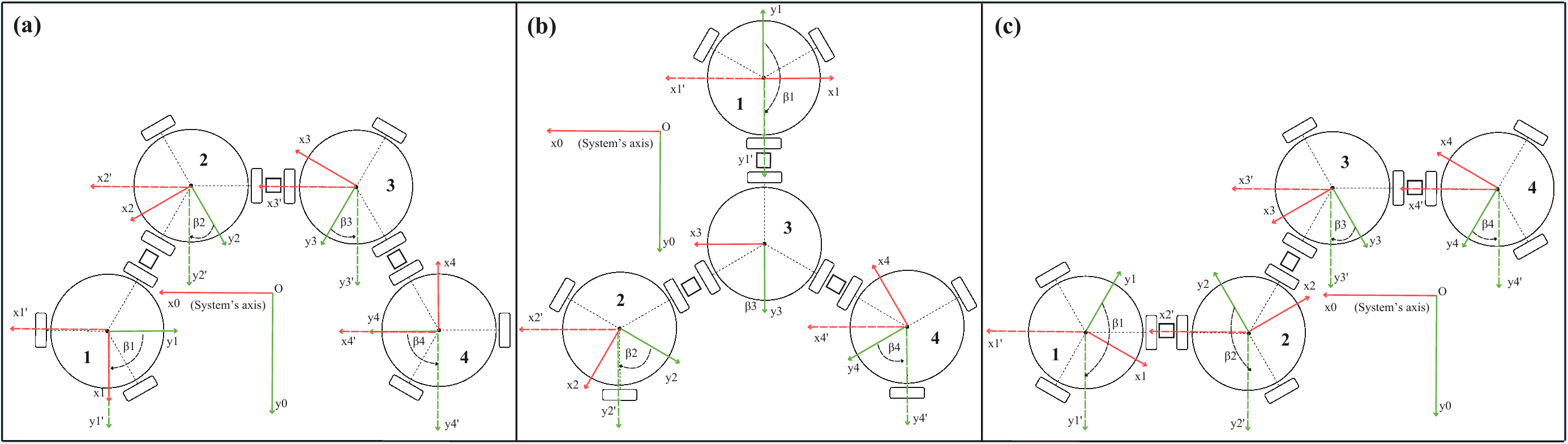}
\vspace{-0.5cm}
\caption{Examples of different docking configurations with 4 mobile manipulators. The solid axes ($Ox_iy_i$) denote the original robot control frames and initial robot arm orientation ($Oy_i$). After docking, the frames are rotated so the dashed axes ($Ox_i'y_i'$) align with the system’s axis, enabling coordinated motion around the new center of mass.}
\vspace{-0.2cm}
\label{fig:Axes}
\end{figure*}

\begin{equation}
\begin{bmatrix}
\dot{\phi}_{1i} \\
\dot{\phi}_{2i} \\
\dot{\phi}_{3i}
\end{bmatrix}
=
\frac{1}{r}
\begin{bmatrix}
-\sin\alpha_1 & \cos\alpha_1 & L \\
-\sin\alpha_2 & \cos\alpha_2 & L \\
-\sin\alpha_3 & \cos\alpha_3 & L
\end{bmatrix}
R(\beta_i)
\begin{bmatrix}
\dot{x_i} \\
\dot{y_i} \\
\dot{\theta_i}
\end{bmatrix}
\end{equation}

where the rotation matrix aligns the local coordinate frame with the unified system axis:

\begin{equation}
R(\beta_i)=
\begin{bmatrix}
\cos\beta_i & -\sin\beta_i & 0 \\
\sin\beta_i & \cos\beta_i & 0 \\
0 & 0 & 1
\end{bmatrix}
\end{equation}

This alignment ensures that all modules move in unison as a rigid platform. Conversely, for rotational control, the wheels located at the docking interfaces are held stationary to serve as rigid pivots, while the remaining peripheral wheels are driven in a coordinated direction to generate a moment around the system's collective center of mass, thereby ensuring stability and preventing internal kinematic conflicts within the docked chain.

\section{Experiments and Evaluations}

This section evaluates the \projectname{} system through four experiments validating our theoretical model. The platform utilizes two LeKiwi modules, each powered by three 12V Feetech ST3215 servos and a Raspberry Pi 5 (8GB) with dual cameras for vision-based alignment. These trials progress from assessing the mechanical reliability of docking (Section~\ref{sec:exp1}) to analyzing post-reconfiguration stability (Section~\ref{sec:4B}, ~\ref{sec:exp3}), finally demonstrating the operational efficiency of the unified bimanual platform in a real-world task (Section~\ref{sec:exp4}).

\subsection{Autonomous Reconfiguration Validation}
\label{sec:exp1}
This experiment identifies the operational boundaries of our vision-based docking strategy, focusing on AprilTag detection limits under varying environmental factors. Establishing this reliability is a prerequisite for all subsequent stability and task-performance trials.

\begin{figure}[t]
\centering
\includegraphics[width=\linewidth]{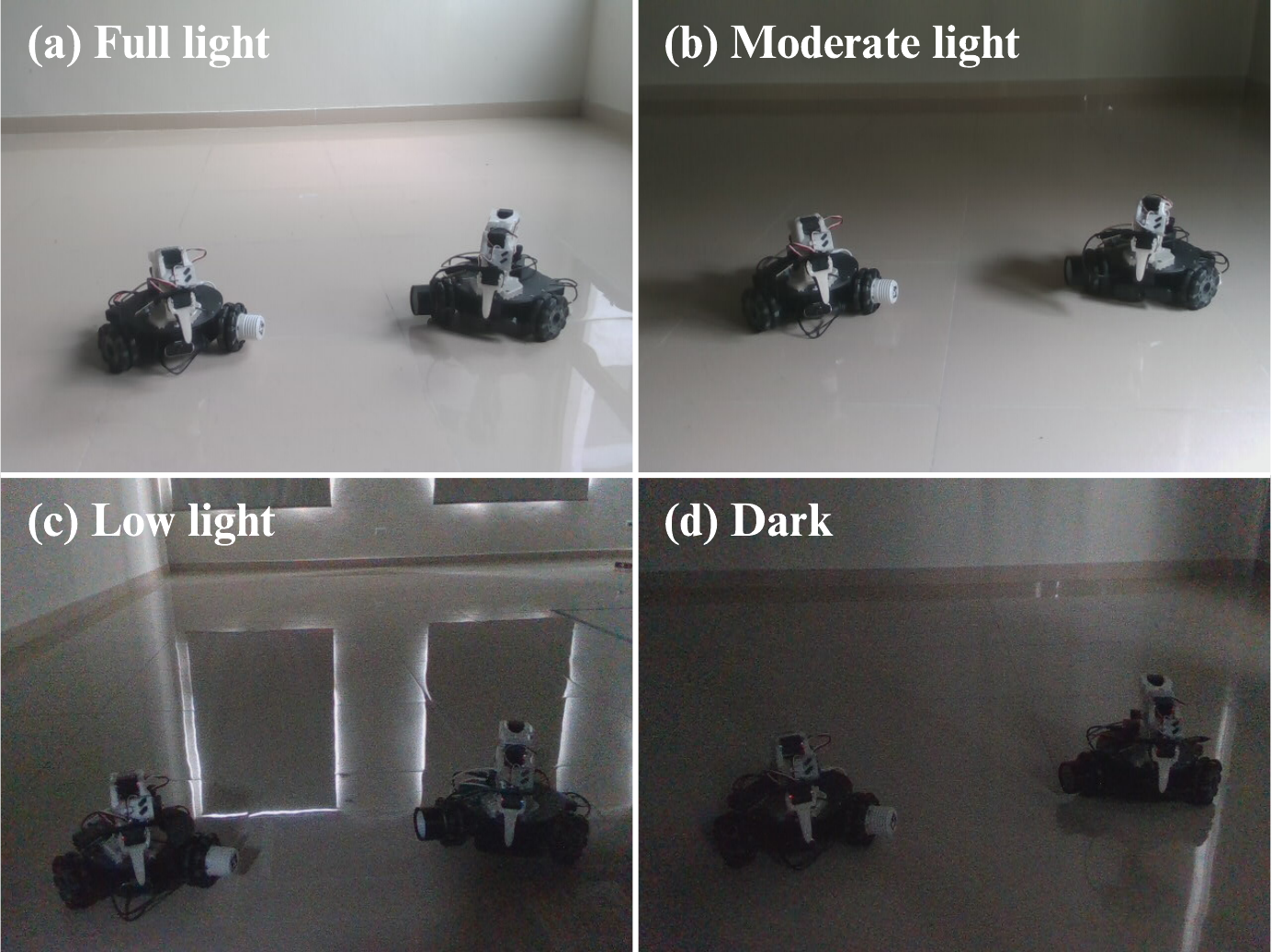}
\caption{The different lighting conditions of the experiment environment used for the docking trials.} 
\vspace{-0.3cm}
\label{fig:lighting_conditions}
\end{figure}

We conducted 15 docking trials across four lighting conditions (Fig. \ref{fig:lighting_conditions}). The system achieved $93.3\%$ success rate, with failures occurring only in total darkness where the camera could not resolve the tag. This demonstrates that the \projectname{} system is robust to most indoor lighting fluctuations.

\begin{figure}[t]
\centering
\includegraphics[width=0.6\linewidth]{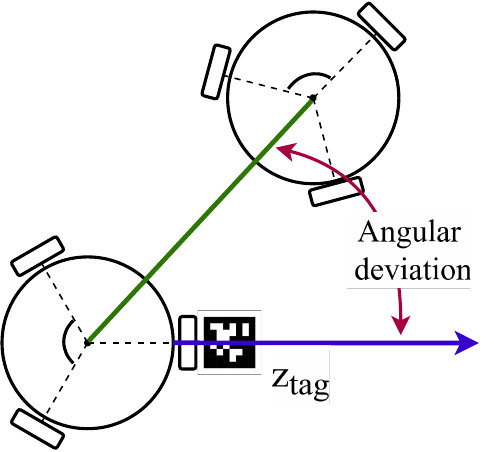}
\caption{Angular deviation, the angle between the AprilTag’s z-axis and the module's connection line, defining the maximum viewing angle for reliable tag detection.}
\vspace{-0.5cm}
\label{fig:tagangle}
\end{figure}

Additionally, we evaluated the detection limit by varying the \textit{angular deviation}—the angle between the AprilTag's z-axis and the modules' line-of-sight (Fig. \ref{fig:tagangle}). Trials revealed a reliable detection threshold of \SI{\pm60}{\degree}, beyond which perspective distortion prevents accurate pose estimation.

Crucially, while the vision system handles the long-range approach, the screw-lock mechanism provides a mechanical self-alignment effect during final contact. This capability to neutralize minor residual errors ensures a successful connection and establishes the zero-backlash foundation required for the high-precision mobility and stability analyzed experiments.

\subsection{Post-Reconfigured Motion Experiment}
\label{sec:4B}
\begin{figure}[h]
\centering
\includegraphics[width=\linewidth]{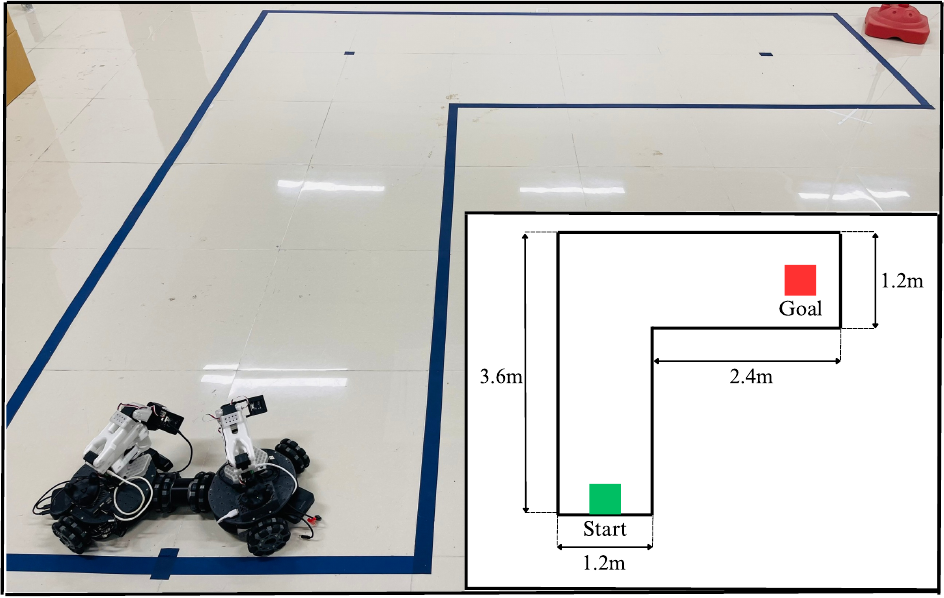}
\caption{Experimental setup for the Post-Reconfigured Motion Experiment, the real-world environment with corresponding dimensional parameters.}
\vspace{-0.5cm}
\label{fig:experimentmap}
\end{figure}

\begin{figure}[h]
\centering
\includegraphics[width=.8\linewidth]{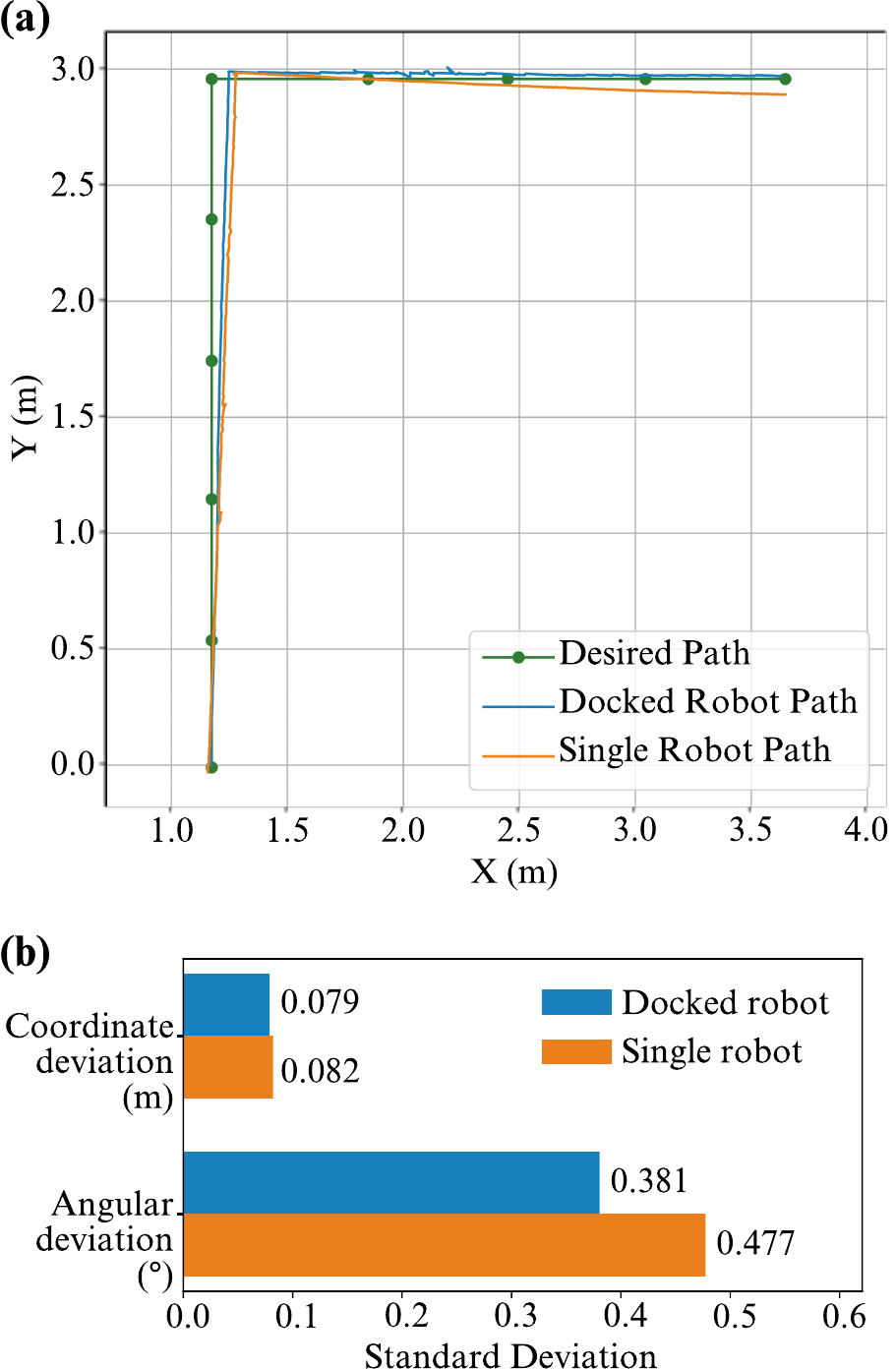}
\caption{Comparison result between the Single and Docked Robot, (a)~visualization of the robot's position and orientation tracking, showing the actual path followed (solid lines) relative to the desired trajectory, (b)~the mean standard deviation of the robot's position and angle provides a quantitative measure of its tracking accuracy.}
\vspace{-0.5cm}
\label{fig:exp2}
\end{figure}

Once the physical link is established, the evaluation shifts to the control stability of the unified kinematic entity. To rigorously quantify performance, the experimental procedure follows a three-stage tracking protocol: (1) reference mapping, where a static Vive Tracker v3.0 establishes a high-resolution L-shaped spatial reference (Fig.~\ref{fig:experimentmap}); (2) real-time feedback control, utilizing an on-robot tracker to minimize instantaneous cross-track and orientation errors $[x,y,\theta]$ at \SI{10}{Hz}; and (3) error cross-referencing between the navigated and desired paths.

Benchmarking the docked system against a single module (Fig.~\ref{fig:exp2}a) confirms the consistency of our centralized control law. Despite coordinating six independent wheels, the docked platform demonstrated kinematic behavior equivalent to a single robot, maintaining a steady heading during forward (\SI{0.1}{m/s}) and lateral (\SI{0.05}{m/s}) maneuvers. This validates that the unified model (Section~\ref{sec:3C}) effectively synchronizes distributed actuators without introducing control lag or path deviation.

Quantitatively, the coordinate standard deviation of the docked system remained comparable to that of a single module (Fig.~\ref{fig:exp2}b). This validates that our control law successfully manages the expanded footprint and actuator redundancy without introducing positional instability. These findings prove that the reconfigured system functions as a reliable, integrated unit, providing the predictable motion baseline necessary for further experiments.

\subsection{Dynamic Stability Experiment}
\label{sec:exp3}
\begin{figure}[h!]
\centering
\includegraphics[width=\linewidth]{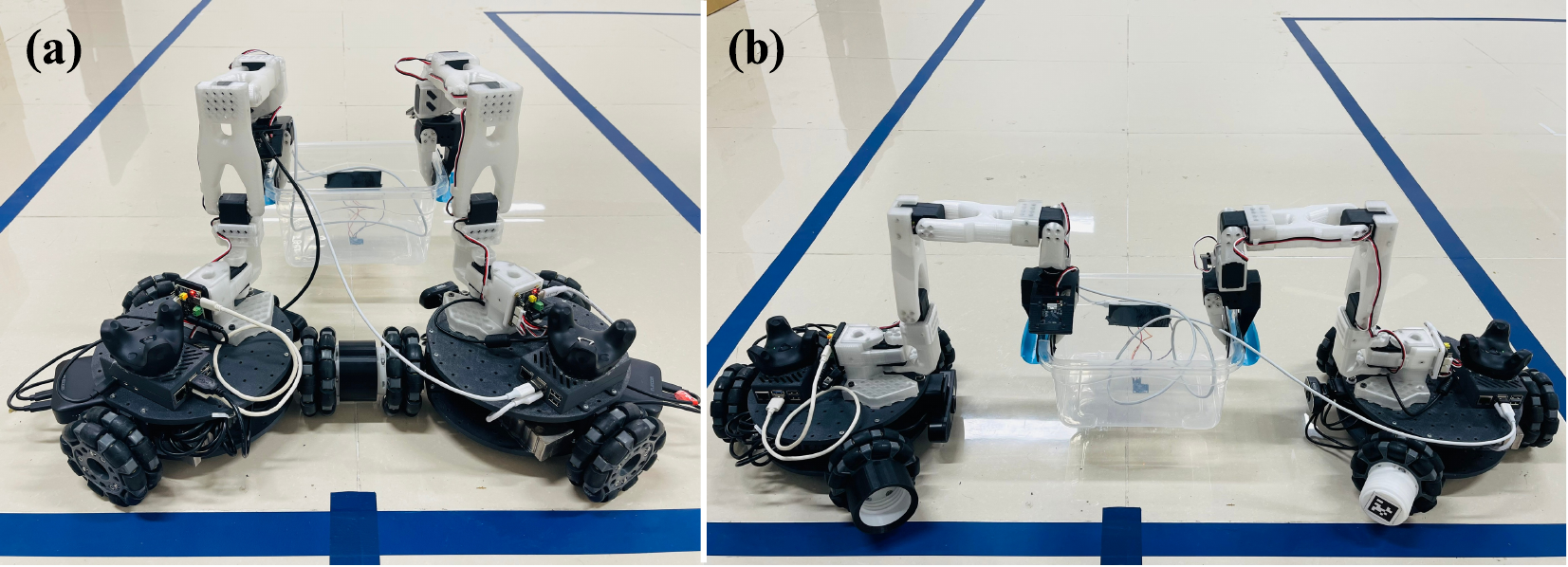}
\caption{Experimental setup for the Stability Experiment, (a) docking robot, (b) cooperation between two independent robots.}
\vspace{-0.5cm}
\label{fig:exp3}
\end{figure}

Based on the experimental setup illustrated in Fig.~\ref{fig:exp3}, we benchmark the unified \projectname{} platform against a dual-robot cooperative team during a heavy-payload task using the L-shaped trajectory from Section~\ref{sec:4B}. Stability was quantified via a multi-modal approach combining high-frequency inertial data and precise spatial tracking. We evaluated performance through four metrics: (1) \textbf{RMSA} (Root Mean Square Acceleration) and (2) \textbf{Jerk} (Derivative of acceleration), derived from the payload-mounted MPU6050 gyroscope to quantify overall motion smoothness and high-frequency perturbations; (3) \textbf{Angular Standard Deviation} ($\sigma_\omega$), extracted from Vive Tracker signals to indicate path-following accuracy; and (4) \textbf{Average Transport Time} for overall operational efficiency.

\begin{figure}[h!]
\centering
\includegraphics[width=.8\linewidth]{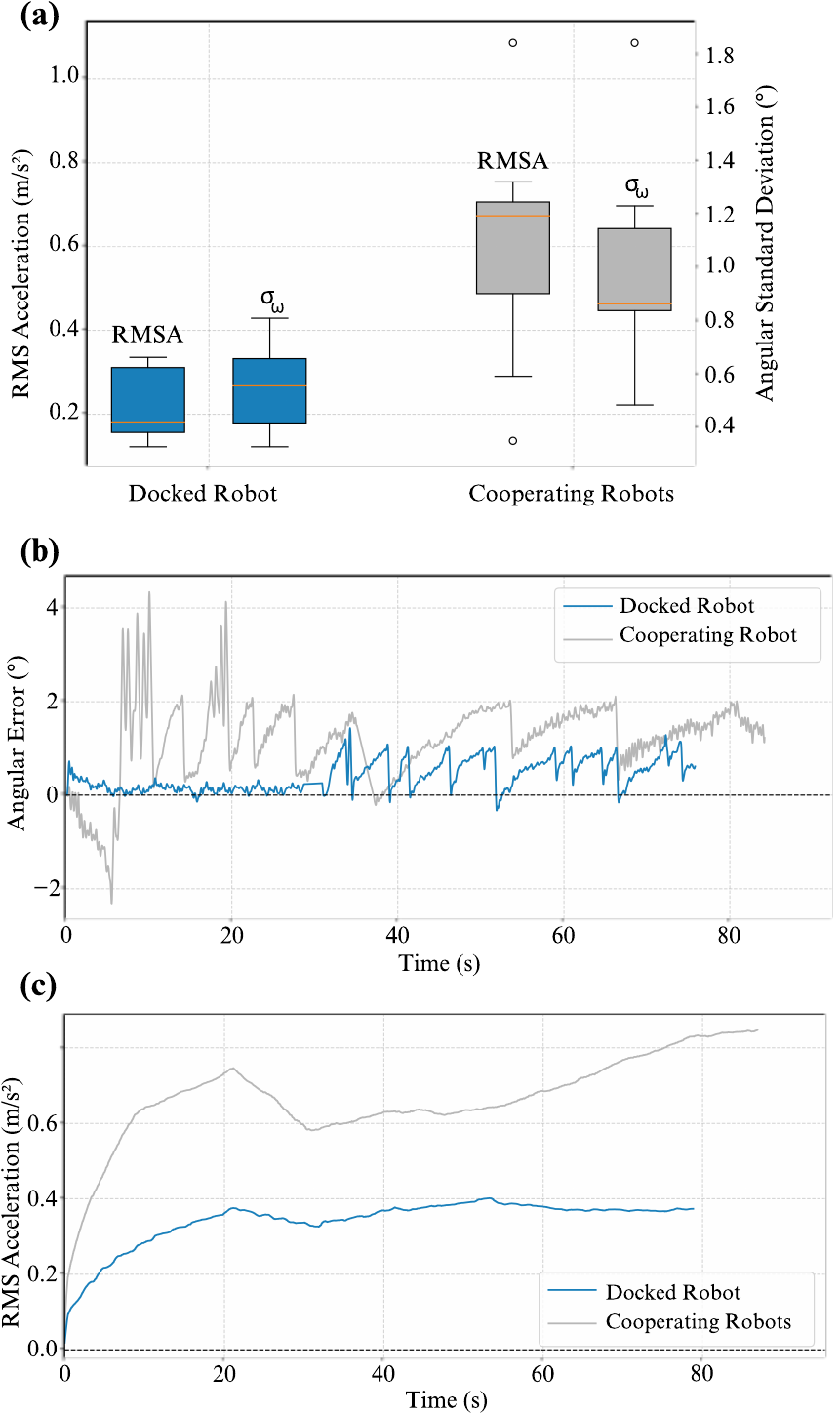}
\caption{Comparative Analysis of Dynamic Stability Docked and Cooperating Robot Systems, (a)~the statistical results of RMS Acceleration and Angular Standard Deviation from 10 runs for the two systems under comparison, (b-c)~the corresponding RMS Acceleration and Angular Standard Deviation results from a single representative run for the two systems.}
\vspace{-0mm}
\label{fig:result3}
\end{figure}

\begin{table}[ht]
\caption{Comparative Analysis of Dynamic Stability and Performance (lower is better)}
\centering
\resizebox{\linewidth}{!}{
\begin{tabular}{lrrrr}
\toprule
\textbf{Mode} & \textbf{RMSA} (\si{\meter\per\second\squared}) & \textbf{Jerk} (\si{\meter\per\second\cubed}) & \textbf{$\sigma_\omega$} (\si{\degree}) & \textbf{Avg. Time} (\si{\second})\\
\midrule
Cooperating & 0.6068 & 2.038 & 1.10 & 84 \\
\textbf{Docking} & \textbf{0.2226} & \textbf{1.710} & \textbf{0.56} & \textbf{76} \\
\bottomrule
\end{tabular}}
\vspace{-0.3cm}
\label{tab:stability_results}
\end{table}


Results averaged over 10 trials (Table~\ref{tab:stability_results}) and visualized in Fig.~\ref{fig:result3} show the docked configuration reduced RMSA and Jerk by \SI{63.3}{\%} and \SI{16.1}{\%}, respectively. This gain is a direct consequence of collapsing the system’s state space from a high-dimensional 9n DOF problem into a single rigid frame. By physically locking the modules, we eliminate redundant degrees of freedom and internal force conflicts that occur when independent actuators "fight" for positioning. Furthermore, the rigid coupling bypasses the communication latencies inherent in wireless coordination, which typically cause asynchronous motor responses. In the docked state, these impulses are absorbed as structural stress, effectively acting as a mechanical low-pass filter. Consequently, the system achieves a \SI{49}{\%} improvement in angular precision ($\sigma_\omega$), proving that \projectname{}'s rigid architecture is superior for payload transport where maximum stability and computational efficiency are paramount.

\subsection{Task Performance Evaluation}
\label{sec:exp4}
\begin{figure}[h!]
\centering
\includegraphics[width=\linewidth]{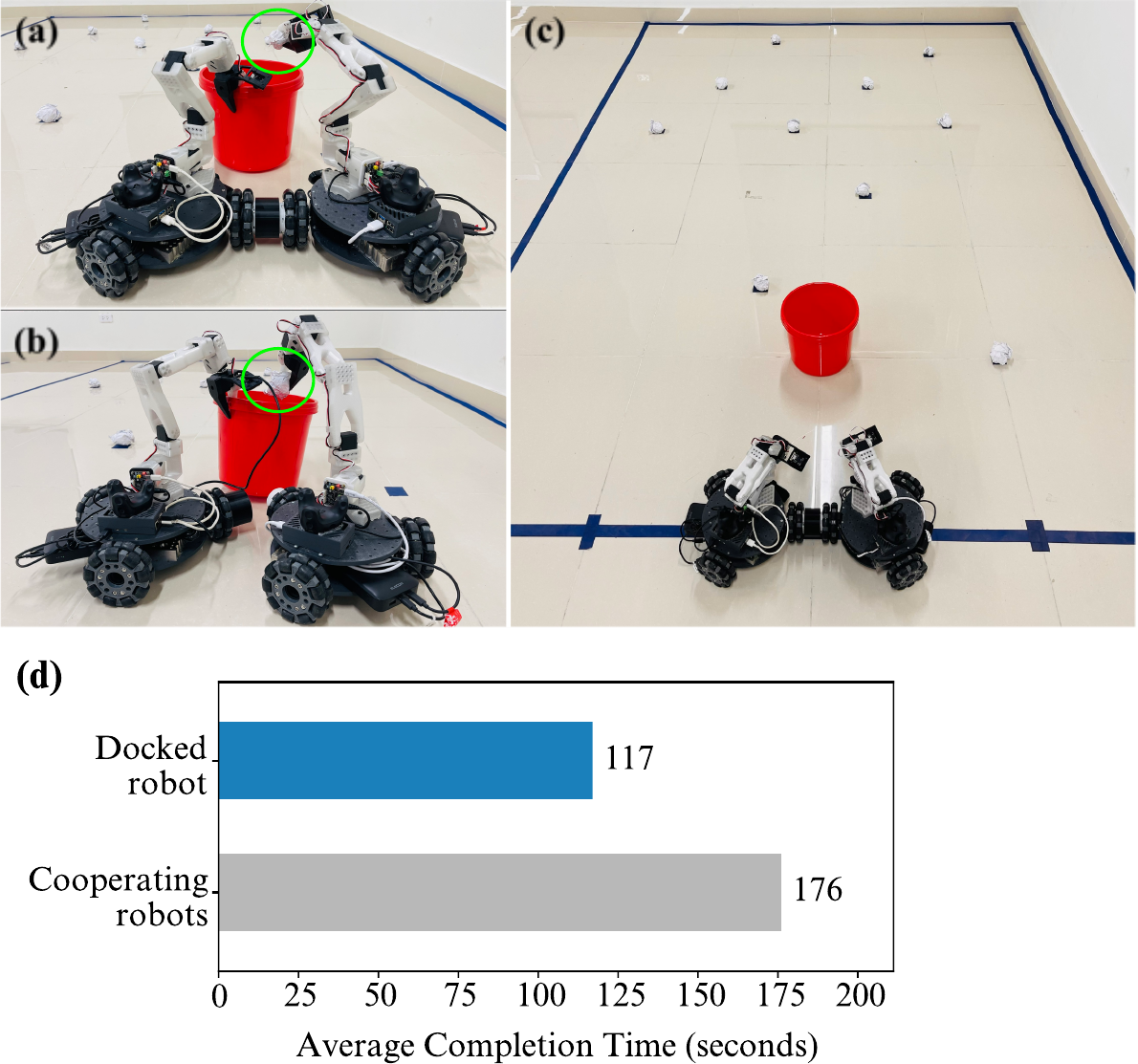}
\caption{Experimental setup for the Task Performance Evaluation, (a)~Docking robot, (b)~Cooperation between two independent robots, (c)~The real-world environment, (d)~Comparative analysis of task completion time for the docked system and two cooperating robots.}
\vspace{-0.25cm}
\label{fig:exp4}
\end{figure}

Physical reconfiguration provides a critical advantage for dual-arm collaborative tasks by transforming a multi-agent coordination problem into a single-base manipulation task. To evaluate this, we conducted a comparative teleoperation study involving two human operators performing a synchronized "trash collection" task. As shown in Fig.~\ref{fig:exp4}(a-c), one module holds a bin while the other picks and places 10 debris items. This task requires continuous spatial synchronization to maintain the relative pose between the manipulator and the receptacle.

The experiment was conducted five times per system within a $\SI{2.4}{\meter} \times \SI{3.6}{\meter}$ workspace. In the cooperative mode, operators must constantly compensate for the independent drift and relative motion between the two mobile bases, leading to high cognitive load and frequent re-alignment pauses. Conversely, the docked system eliminates relative base motion, allowing operators to focus exclusively on end-effector precision.

As illustrated in Fig.~\ref{fig:exp4}d, the docked configuration achieved a significant reduction in execution time. These results confirm that physical docking simplifies the control strategy by removing the need for active base-to-base synchronization. Furthermore, this success highlights a broader potential: the docking paradigm can effectively resolve complex manipulation tasks that are otherwise bottlenecked by the coordination overhead and communication constraints of multi-robot systems, offering a more robust and scalable alternative for high-precision collaborative robotics.

\section{DISCUSSION \& FUTURE WORKS}

The experimental results demonstrate that physical reconfiguration via \projectname{} offers significant advantages in stability and control simplicity. However, it is important to clarify that this docking paradigm is intended to complement, not replace, traditional multi-robot coordination. Cooperative systems remain essential for tasks requiring high flexibility, such as transporting oversized objects or navigating environments where modules must remain spatially separated. Our proposed system introduces a versatile hybrid morphology, allowing a multi-robot team to switch between three operational modes: independent, cooperative, and docked. This multi-modal capability significantly expands the functional scope of mobile manipulators in dynamic environments.

Despite these strengths, certain limitations persist. The current docking and alignment mechanism is optimized for planar 2D surfaces. In its present form, uneven terrain could hinder the precision of the screw-lock engagement. Additionally, the reliance on teleoperation for task evaluation introduces operator-dependent variability.

In future work, we will focus on:
\begin{itemize}
\item 3D Docking Mechanisms: Developing advanced alignment strategies to support reconfiguration on non-planar surfaces and multi-level environments.
\item Autonomous Coordination: Implementing Reinforcement Learning (RL) to transition from teleoperation to fully autonomous task execution, ensuring more objective performance benchmarks.
\item System Scaling: Researching control strategies for scaling beyond two modules, focusing on force distribution and communication protocols for large-scale reconfigurable swarms.
\end{itemize}

\section{CONCLUSIONS}

In this research, we introduced \projectname{}, a modular reconfigurable mobile manipulator system designed to bridge the gap between multi-robot flexibility and single-platform stability. We presented an integrated framework encompassing an innovative mechanical docking design, a vision-based alignment strategy, and a unified control law. Our experiments validate that the docked configuration achieves kinematic equivalence to a single robot while providing superior dynamic stability (reduced RMSA and Jerk) and operational efficiency compared to independent cooperative teams.

The findings confirm that establishing a rigid physical link effectively collapses the control complexity and eliminates the synchronization bottlenecks typical of multi-agent systems. By offering a new morphological ``mode" for heavy-duty and high-precision tasks, \projectname{} provides a robust foundation for the next generation of collaborative robots in complex, real-world applications.









\addtolength{\textheight}{-6cm}   
\bibliographystyle{IEEEtran}
\bibliography{biblio}

\end{document}